\documentclass[10pt,twocolumn]{article}

\usepackage[margin=0.72in,columnsep=0.24in]{geometry}
\usepackage[T1]{fontenc}
\usepackage{lmodern}
\usepackage{tgtermes} 
\usepackage{microtype}
\usepackage{graphicx}
\usepackage{booktabs}
\usepackage{array}
\usepackage{tabularx}
\usepackage{xcolor}
\usepackage{hyperref}
\usepackage[numbers,sort&compress]{natbib}
\usepackage{enumitem}
\usepackage{caption}
\usepackage{float}
\usepackage{amsmath}
\usepackage{xspace}
\usepackage{fancyhdr}

\hypersetup{colorlinks=true,allcolors=blue!55!black}
\setlist{nosep,leftmargin=*}
\fancypagestyle{plain}{%
  \fancyhf{}%
  \fancyfoot[C]{\footnotesize Preprint}%
  \fancyfoot[R]{\footnotesize\thepage}%
}
\makeatletter
\renewcommand\@seccntformat[1]{\csname the#1\endcsname\hspace{0.65em}}
\renewcommand\subsection{%
  \@startsection{subsection}{2}{\z@}%
    {-1.25em \@plus -0.25em \@minus -0.1em}%
    {0.55em \@plus 0.1em}%
    {\normalfont\fontsize{10.5}{12.5}\selectfont\bfseries\raggedright
     \hyphenpenalty=10000\exhyphenpenalty=10000}}
\makeatother
\newcommand{\system}{\textsf{ConversationalVoice}\xspace}
\newcommand{\stage}[1]{\textsf{#1}}

\title{%
  \vspace{-1.0em}
  {\fontsize{22}{25}\selectfont\bfseries ConversationalVoice}\\[0.28em]
  {\fontsize{17.5}{21}\selectfont\bfseries Full-Duplex Speech Data from Real Conversations}\\[0.12em]
  {\fontsize{15.5}{18.5}\selectfont\bfseries through Source-Faithful Reconstruction}\\[-0.04em]
  {\fontsize{15.5}{18.5}\selectfont\bfseries and Conversation-Grounded Expansion}
}
\author{%
  Richard Yucheng He, Baodong Cao, Chen Xu, Yihang Liu, Tairan Chen\\[0.65em]
  {\large\bfseries AveraLabs}\\[0.2em]
  {\small\ttfamily\{richard, baodong.cao, chen.xu, robbie.liu,
  terrence.chen\}@averalabs.com}%
}
\date{}

\begin{document}
\maketitle

\begin{abstract}
Full-duplex speech models require training data that preserves turn-taking, overlap, interruption, and backchannel behavior, yet these signals are entangled across speakers in noisy real-world recordings. We present \system, a pipeline that converts real two-speaker excerpts into three complementary training-data artifacts. (1) Separation recovers speaker-specific tracks with stable speaker assignments, a canonical transcript, and naturally observed interaction timing. (2) Reconstruction generates speech in matched voices from a fixed source transcript, reconstructs the source turn order, pauses, and overlaps, and adds word-level alignment and delivery instructions. (3) Expansion generates new dialogue constrained by the source context, speakers, and observed interaction pattern. Automatic speaker-verification metrics remain strong across stages, with same-speaker similarity of 0.983--0.991 and positive discrimination margins of 0.199--0.209. Predicted speech quality (NISQA MOS) is 3.56 for separation, 4.41 for reconstruction, and 4.61 for expansion. A Gemini-based automatic evaluator assigns expansion mean scores of 4.94/5 for contextual coherence and 4.80/5 for dialogue naturalness. Expansion and reconstruction exhibit broadly similar interaction profiles; expansion's turn, overlap-event, backchannel, and interruption rates are 4.6\%, 8.0\%, 13.2\%, and 16.0\% lower, respectively. We evaluate data properties only; downstream gains in full-duplex model training remain for future work.
\end{abstract}

{\raggedright
\noindent\textbf{Code repository:}\\[-0.1em]
{\small\url{https://github.com/avera-labs/ConversationalVoice}}\par
}


\begin{figure*}[t]
    \centering
    \includegraphics[width=0.96\textwidth]{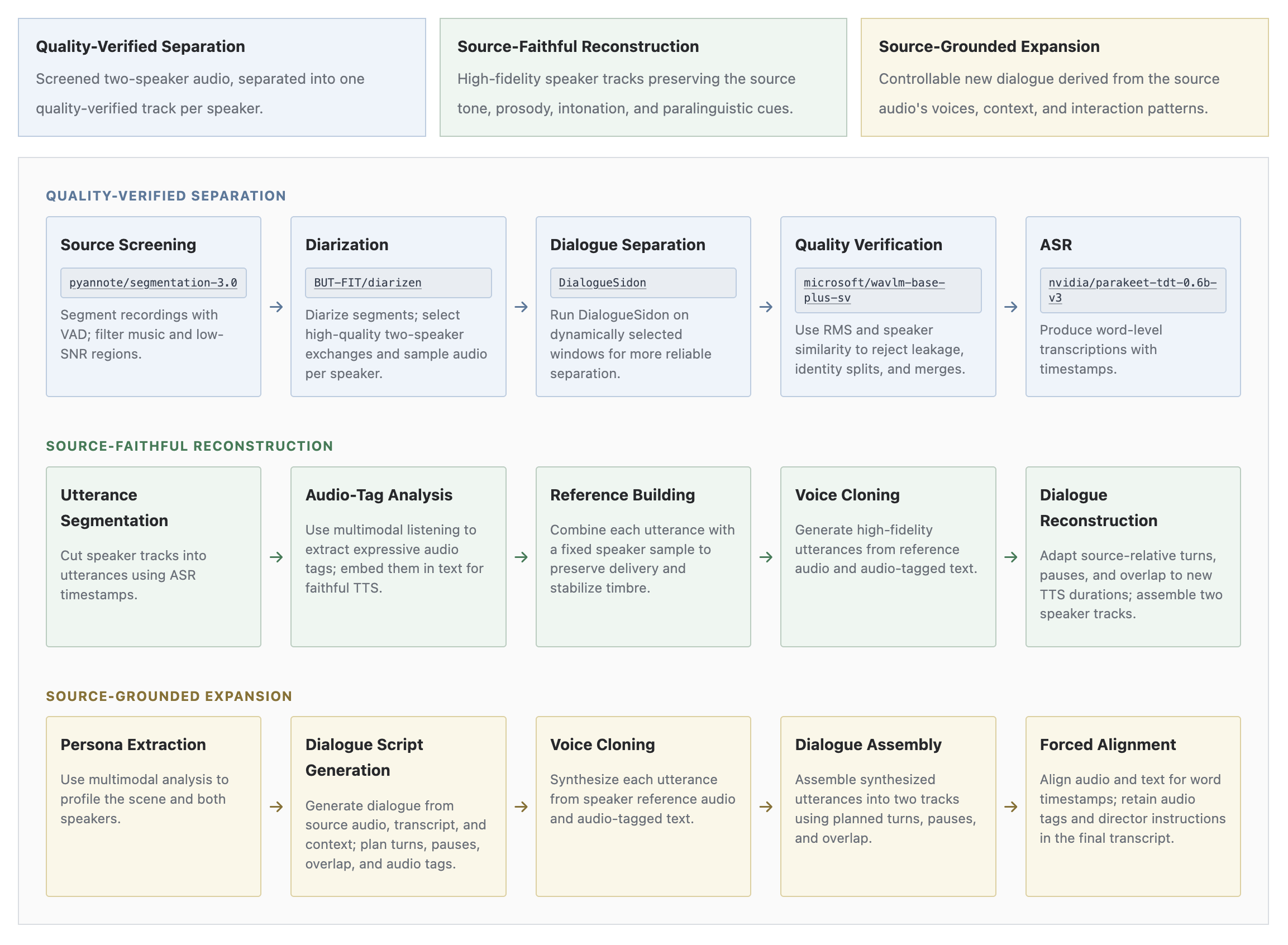}
    \caption{\system produces three complementary views of a validated conversation. Separation recovers speakers; reconstruction regenerates the source exchange; expansion creates a grounded continuation. Reconstruction and expansion additionally expose word timing and delivery-instruction metadata.}
    \label{fig:pipeline}
\end{figure*}

\section{Introduction}

Training full-duplex speech models involves modeling conversations as two synchronized speaker streams rather than as isolated utterances. Accordingly, training data for these models should preserve words, speaker identity, turn transitions, silence, overlap, feedback, and nonverbal behavior. Full-duplex speech models increasingly represent these temporal signals explicitly, while related benchmarks systematically evaluate them \citep{nguyen2022dgslm,defossez2024moshi,lin2025fullbench}; yet real conversations usually arrive as noisy monaural recordings, while prompt-generated dialogue can lose the timing and interaction patterns of actual exchanges.

We present \system, an end-to-end pipeline that transforms real-world two-speaker audio excerpts into three complementary forms of full-duplex training data. \emph{Quality-verified separation} combines diarization, dialogue separation, stable speaker-slot assignment, signal and identity checks, and ASR to recover speaker-specific tracks with a canonical word-level transcript. Its contribution is to make separation a validated substrate with fixed speaker-slot assignments for subsequent generation rather than the final dataset.

\emph{Source-faithful reconstruction} regenerates the recovered exchange while fixing its words and speaker identities. Stable speaker references and utterance-local acoustic cues guide synthesis; forced alignment and conversation-level scheduling then rebuild word timing, turns, pauses, and overlaps around the generated durations.

\emph{Conversation-grounded expansion} generates new dialogue under constraints from the source transcript, audio, speaker profiles, and observed interaction. It is designed to add conversational coverage while explicitly planning dialogue, backchannels, paralinguistic events, and sequential or overlapping placement.

Both generated stages produce paired, time-aligned single-speaker tracks with speaker-attributed transcripts, word-level timestamps, audio tags and delivery instructions. The three stages therefore recover, reconstruct, and extend real conversational evidence in a shared training representation.

Specifically, this paper makes four contributions:
\begin{enumerate}
    \item We operationalize separation as a quality-verified data stage that maintains stable speaker-slot assignments and supplies canonical transcription for downstream transformations.
    \item We introduce reconstruction that preserves source words and speaker-to-track assignments while regenerating clean tracks and adapting conversational timing to synthesized speech.
    \item We introduce expansion that creates new interactions grounded in real speakers, context, and explicitly represented turn-taking and paralinguistic behavior.
    \item We define a shared training artifact with aligned speaker tracks, word timing, audio tags, delivery instructions.
\end{enumerate}

The automatic evaluation characterizes shared output quality, reconstruction fidelity, and expansion interaction and content quality. It tests whether the three artifacts retain the properties needed for their complementary roles; downstream model gains remain outside the scope of this work.

\section{Related Work}

\subsection{Full-duplex spoken dialogue modeling}

Spoken dialogue models have moved from turn-concatenated utterances toward synchronized streams. The dialogue generative spoken language model (dGSLM) learns from two-channel conversational speech and generates speech conditionally across channels \citep{nguyen2022dgslm}. Moshi models user and system audio in parallel and uses time-aligned text as an intermediate prediction signal, illustrating the value of preserving both timing and lexical structure \citep{defossez2024moshi}. Full-Duplex-Bench organizes evaluation around turn-taking abilities such as user pause, backchannel, interruption, and simultaneous speech \citep{lin2025fullbench}. Together, these systems motivate training examples that retain independent channels and fine temporal structure rather than collapsing a conversation into alternating text turns.

\subsection{Recovering conversation from in-the-wild audio}

Speech separation estimates multiple source signals from a mixture. SepFormer demonstrated the effectiveness of attention-based temporal modeling for clean separation benchmarks \citep{subakan2021sepformer}. In-the-wild dialogue adds degradations not well represented by clean synthetic mixtures. DialogueSidon addresses this setting through joint restoration and separation of degraded monaural two-speaker audio \citep{nakata2026dialoguesidon}. DuplexChat builds on this capability in a data pipeline that filters public podcast audio, identifies two-speaker clips, and creates separated full-duplex tracks at large scale \citep{nakata2026duplexchat}.

DuplexChat is the closest prior system to the first stage of \system. Its principal endpoint is a speaker-separated corpus that preserves naturally occurring turn-taking. Building on DuplexChat, we strengthen the separation stage through screening, stable speaker assignment, quality verification, and canonical transcription, and then extend its outputs through source-faithful reconstruction and conversation-grounded expansion. Reconstruction converts the separated content into cleaner, source-faithful synthesized speech with explicit fine-grained supervision. Expansion uses the recovered exchange to produce new but grounded interactions. The distinction is functional: separation recovers streams already present in the recording; reconstruction regenerates the same linguistic event; expansion adds a new conversational event constrained by the original context.

\subsection{Expressive speech generation and rich annotation}

Modern TTS systems provide multilingual synthesis, voice cloning, and instruction-based control. Qwen3-TTS, for example, reports short-reference voice cloning and description-based control in a multilingual architecture \citep{hu2026qwen3tts}. SoulX-Podcast targets long-form multi-speaker dialogue and incorporates dialectal and paralinguistic controls \citep{xie2025soulx}. These advances make controlled resynthesis and conversation generation practical, but a TTS model does not determine the data semantics. Those semantics remain pipeline-level design choices: which content is immutable, how identity is anchored, how overlaps are scheduled, and how delivery instructions are exposed as supervision.

Dataset work also shows the value of retaining information beyond words. NaturalVoices derives spontaneous podcast speech with annotations for emotion, speech quality, transcripts, speaker identity, and sound events \citep{du2025naturalvoices}. WavLM supplies representations useful for speaker recognition and other non-ASR tasks \citep{chen2022wavlm}. \system combines these concerns in a single conversation-level artifact. The objective is not to introduce a new separator, TTS model, or speaker encoder, but to leverage the complementary strengths of existing components to construct full-duplex speech training data from real dialogue.

\section{ConversationalVoice Pipeline}

Figure~\ref{fig:pipeline} summarizes the pipeline. The three stages share speaker identities and source provenance but serve different training purposes. Separation retains maximum fidelity to the observed waveform. Reconstruction keeps the observed words and interaction while improving controllability and acoustic cleanliness. Expansion is intended to increase conversational coverage while remaining anchored to the source scene and speakers.

\subsection{Screening, diarization, and two-speaker selection}

The input is an arbitrary real-world recording expected to contain conversational speech. Audio is normalized to 16~kHz mono for analysis. Voice activity detection proposes speech regions, after which music and low-signal-to-noise regions are filtered. Diarization assigns speaker hypotheses over time. The planner retains windows that satisfy a strict two-speaker structure and selects clean reference excerpts for both speakers.

Window selection is dynamic rather than a fixed-duration split. A candidate should contain enough speech from both speakers to identify them, but avoid boundaries that truncate an exchange or make separation unnecessarily difficult. The planner therefore uses diarization boundaries and activity statistics to choose context windows. This is important for real dialogue: an overlap may be short relative to the surrounding exchange, and a fixed cut can remove the context required to map separated output slots back to stable identities.

\subsection{Quality-verified separation and transcription}

Each accepted window is processed by DialogueSidon, which jointly separates and restores degraded two-speaker dialogue \citep{nakata2026dialoguesidon}. The separator returns two output slots, but slot order has no intrinsic identity. \system maps each slot to a diarization speaker using activity correspondence and speaker embeddings, then keeps that mapping fixed for the conversation. This prevents a local permutation from becoming an identity switch in downstream tracks.

Quality verification combines signal activity and speaker consistency. Silent or implausibly weak tracks are rejected. WavLM speaker-verification embeddings \citep{chen2022wavlm} compare candidate regions against canonical speaker references to detect leakage, merges, splits, and mismatched slots. A failed window can be retried under a revised selection; persistent failures are excluded rather than propagated to synthesis.

Language-specific ASR produces a speaker-attributed transcript. The transcript establishes utterance text, source boundaries, and an initial word timeline. This canonical transcript serves as the fixed lexical reference for reconstruction and the primary semantic context for expansion.

\subsection{Source-faithful reconstruction}

Reconstruction asks a constrained question: given what a person actually said and how the exchange unfolded, can the pipeline produce a cleaner single-speaker rendering while keeping the lexical event and speaker identity intact? The process operates utterance by utterance.

For each utterance, the corresponding region is sliced from the separated speaker track. A multimodal analysis step listens to the slice and produces two fields. The first, \texttt{text\_with\_audio\_tags}, augments the canonical transcript with tags from an approved taxonomy. Tags are position preserving: deleting them recovers the canonical source text exactly. The second field is a single actor-facing \texttt{instruction} that describes delivery without changing what is spoken. The validation layer rejects substitutions, additions, speaker labels, or stage directions that leak into lexical content.

Voice generation uses a composite reference. A fixed clean sample for the speaker establishes global identity. It is followed by exactly one second of silence and the separated source utterance, which supplies local delivery evidence such as tempo, prosody, and nonverbal behavior. The fixed component reduces identity drift between utterances; the local component avoids asking a text-only controller to infer every expressive detail. The tagged text, instruction, and composite reference are passed to a voice-cloning TTS service to synthesize an isolated waveform.

Word alignment is computed on each isolated synthesized utterance before it is placed on the conversation timeline. This ordering is deliberate: we observed timestamp drift when forced alignment was applied after assembly, particularly at boundaries between speech and long silences; aligning isolated utterances before inserting silence avoids this source of error.

The global timeline cannot simply reuse every source boundary because TTS duration differs from source duration. \system therefore reconstructs relative interaction. It preserves turn order and source gap intent, then adapts pause and overlap placement to the synthesized durations. If the next utterance originally begins after the current one ends, the corresponding nonnegative gap is retained. If it begins before the current one ends, the scheduler preserves an overlap relation while respecting the generated duration. A speaker is never scheduled to overlap with itself. Finally, local word times are shifted by the scheduled utterance onset.

The output contains two mono PCM16 waveforms at 44.1~kHz, padded to equal duration, plus a transcript and manifest. One track contains only speaker~0 events and the other only speaker~1 events. The transcript preserves both the plain source text and its tagged form, which allows a training pipeline to choose lexical-only, tagged, or instruction-conditioned objectives without reprocessing the waveform.

\subsection{Conversation-grounded expansion}

Expansion generates a continuation rather than a paraphrase of the source chunk. Its context includes the canonical transcript, source audio, speaker profiles, stable reference recordings, and a summary of the preceding exchange. The dialogue generator is instructed to maintain the established participants and situation while creating new content. Source provenance is retained so that an expansion can always be traced to the conversation that grounded it.

The script schema distinguishes \texttt{dialogue}, \texttt{backchannel}, and \texttt{paralinguistic} utterances. It also specifies whether an event is sequential or overlaps an active event. These fields turn interaction structure into an explicit generation plan. Tagged text expresses audible events, while the instruction field stores the delivery instruction. Plain text is derived from tagged text by the worker rather than independently generated, reducing disagreement between the two representations.

Before synthesis, schema and semantic checks enforce valid speakers, ordered events, allowed tags, and content consistency. Each utterance is then generated with the matching speaker reference and forced aligned. The assembler places waveforms according to the planned turns, pauses, and overlaps, prevents self-overlap, and produces two equal-duration tracks. Expansion has its own time origin and artifact identity; it is not silently concatenated to the source recording. This makes duration accounting and lineage unambiguous.

Grounding constrains expansion without requiring acoustic imitation of every source event. The continuation may introduce new words and new turn sequences, but it should remain compatible with the people, context, and interaction style established by the source. This middle ground separates it from source reconstruction on one side and unconstrained prompt-generated dialogue on the other.

\subsection{Training artifact and provenance}

Table~\ref{tab:schema} summarizes the transcript unit. Word items and audio-tag items share a timeline. A word has a nonzero interval estimated by forced alignment. An audio tag is represented as a zero-duration anchor at the point where the event is intended or detected. This avoids assigning an arbitrary lexical duration to a laugh, breath, or similar event while preserving its order relative to words.

\begin{table}[t]
\centering
\caption{Core fields in a generated utterance artifact.}
\label{tab:schema}
\small
\begin{tabularx}{\columnwidth}{@{}p{0.28\columnwidth}X@{}}
\toprule
Field & Role \\
\midrule
\texttt{speaker} & Stable identity and output-track assignment \\
\texttt{utterance\_type} & Dialogue, backchannel, or paralinguistic event \\
\texttt{start}, \texttt{end} & Conversation-level utterance interval \\
\texttt{text} & Plain lexical content \\
\shortstack[l]{\texttt{text\_with\_}\\\texttt{audio\_tags}} & Position-preserving expressive annotation \\
\texttt{instruction} & Actor-facing delivery instruction \\
\texttt{words} & Word intervals and zero-duration tag anchors \\
\bottomrule
\end{tabularx}
\end{table}

The manifest records input identity, stage, model identifiers, configuration, artifact locations, sizes, and hashes. These fields separate semantic provenance from storage location and allow a later evaluation run to identify exactly which inputs and generated outputs were scored. Model revisions and service behavior can change, so a reproducible release should freeze the resolved model revisions and configuration snapshot rather than only record a mutable product name.

\section{Evaluation}

\subsection{Evaluation protocol and score groups}

\system produces usable speaker-separated data at three stages: \stage{separation}, \stage{reconstruction}, and \stage{expansion}. We evaluate and compare these three outputs using five complementary score groups. \textbf{Shared Output Quality} evaluates script adherence, predicted acoustic quality, and speaker identity for \stage{separation}, \stage{reconstruction}, and \stage{expansion}. \textbf{Reconstruction Fidelity} evaluates whether reconstruction preserves the duration and interaction structure of its paired separation source. \textbf{Expansion Statistics and Analysis} compares the duration and interaction density of a generated continuation with its paired reconstruction. \textbf{Expansion Content and Dialogue Quality} evaluates whether the continuation is coherent with that reconstruction and forms a natural two-person conversation. \textbf{Audio-tag Annotation Quality} separately evaluates how faithfully declared tags are expressed in generated utterances.

WER uses the remotely hosted \texttt{qwen3-asr-1.7b} model; NISQA, DNSMOS, WavLM speaker embeddings, VAD, event detection, and aggregation use frozen local implementations and weights. ASR, acoustic-quality, and speaker-identity metrics operate on transcript-derived active speech to prevent scheduled silence from dominating the estimates. Duration and interaction metrics use the effective conversation interval, defined from the first detected speech onset to the last speech offset across both tracks, thereby excluding terminal padding. Shared-quality metrics and duration ratios use equal weighting across evaluation units, interaction rates pool event counts over effective conversation duration, and Audio-tag Alignment is averaged across successfully evaluated tagged utterances.

\subsection{Shared Output Quality}

These metrics assess whether each stage produces usable speech without requiring waveform correspondence to another stage:
\begin{itemize}
    \item \textbf{WER:} We mix the two output tracks and transcribe the result with \texttt{qwen3-asr-1.7b}. The reference is constructed by concatenating the plain utterance text in chronological order after removing audio tags. For separation, the reference is the canonical speaker-attributed transcript produced by the transcription stage. For reconstruction, it is the reconstruction transcript, whose lexical content is copied unchanged from the canonical transcript while its timing and annotations are updated for the synthesized output. For expansion, it is the generated continuation script. WER is computed as $(S+D+I)/N_{\mathrm{ref}}$ after English text normalization.
    \item \textbf{NISQA:} NISQA predicts overall no-reference speech quality \citep{mittag2021nisqa}. Active speech is segmented into windows of at most 50 seconds, and window scores are aggregated by duration. Noisiness, coloration, discontinuity, and loudness are retained as diagnostic dimensions.
    \item \textbf{DNSMOS:} DNSMOS provides a complementary non-intrusive quality estimate \citep{reddy2021dnsmos}. We report OVRL as the summary score and retain SIG, BAK, and P808 for diagnosis.
    \item \textbf{Speaker identity:} For each output track, Same-speaker Similarity is the cosine similarity between its duration-pooled WavLM speaker-verification embedding and the assigned canonical reference \citep{chen2022wavlm}. Speaker Discrimination Margin is $s_{\mathrm{assigned}}-s_{\mathrm{other}}$. A positive value means that the track-level embedding is closer to the assigned speaker than to the other participant.
\end{itemize}

\begin{table*}[t]
\centering
\caption{Shared Output Quality. WER is computed against the stage-specific reference transcripts defined in the text and is lower-is-better; all other metrics are higher-is-better.}
\label{tab:results}
\small
\begin{tabular}{lccccc}
\toprule
Stage & WER & NISQA MOS & DNSMOS OVRL & Same-speaker & Margin \\
\midrule
Separation & 0.140 & 3.563 & 3.287 & 0.983 & 0.200 \\
Reconstruction & 0.150 & 4.405 & 3.400 & 0.988 & 0.199 \\
Expansion & 0.039 & 4.608 & 3.328 & 0.991 & 0.209 \\
\bottomrule
\end{tabular}
\end{table*}

Table~\ref{tab:results} shows a positive speaker-discrimination margin at every stage. Relative to separation, NISQA MOS increases from 3.563 to 4.405 for reconstruction and 4.608 for expansion; DNSMOS OVRL changes from 3.287 to 3.400 and 3.328, respectively. Because the reconstruction transcript retains the canonical source words, separation and reconstruction are evaluated against the same lexical content, although each stage uses its own time-aligned transcript artifact. Reconstruction WER is 0.150, compared with 0.140 for separation. Expansion obtains a WER of 0.039 against its own generated continuation script; this lower value measures script realization and should not be interpreted as a cross-stage improvement in transcription or semantic accuracy. Same-speaker Similarity ranges from 0.983 to 0.991, with positive mean Speaker Discrimination Margins across all stages. Together with the acoustic-quality scores, these results indicate strong predicted speech quality and high aggregate similarity to the assigned speaker references.

\subsection{Reconstruction Fidelity}

Reconstruction replaces the acoustic realization while retaining the linguistic and conversational content of the separated dialogue. Fidelity is therefore evaluated against the paired separation source using:
\begin{itemize}
    \item \textbf{Duration Ratio:} $D_{\mathrm{rec}}/D_{\mathrm{sep}}$, with a target of one.
    \item \textbf{Event preservation:} Turn, overlap, and backchannel events are detected with a frozen VAD configuration and matched one-to-one using the corresponding source utterances rather than their absolute timestamps. Turn matching requires speaker and source-utterance agreement while preserving order. Overlap matching requires the same source-utterance pair, after merging VAD fragments from that pair separated by at most 300~ms. Backchannel matching requires the same feedback utterance and conversational anchor. We compute $F_1=2TP/(2TP+FP+FN)$. When reference or predicted events exist but no pair matches, the valid score is zero.
\end{itemize}

\begin{table*}[t]
\centering
\caption{Reconstruction Fidelity against the paired separation source. Duration Ratio targets one; F1 metrics are higher-is-better.}
\label{tab:reconstruction}
\small
\begin{tabular}{cccc}
\toprule
Duration Ratio & Turn $F_1$ & Overlap $F_1$ & Backchannel $F_1$ \\
\midrule
1.263 & 0.693 & 0.664 & 0.860 \\
\bottomrule
\end{tabular}
\end{table*}

Table~\ref{tab:reconstruction} shows that reconstruction is 26.3\% longer than its paired separation source on average, reflecting the duration change introduced by resynthesis. Under source-utterance-relative matching, Turn and Overlap preservation reach $F_1=0.693$ and $0.664$, respectively, while Backchannel preservation reaches $F_1=0.860$. The generated waveform therefore differs in absolute duration while retaining substantial correspondence to the source interaction structure.

\subsection{Expansion Statistics and Analysis}

Expansion creates a new continuation, so reconstruction events are contextual references rather than one-to-one targets. We therefore compare descriptive interaction statistics over the evaluated reconstruction--expansion pairs:
\begin{itemize}
    \item \textbf{Expansion Factor:} $D_{\mathrm{exp}}/D_{\mathrm{rec}}$, a descriptive measure of generated duration rather than a monotonic quality score.
    \item \textbf{Interaction rates:} Turn, backchannel, interruption, and distinct overlap-event counts divided by effective conversation minutes. A distinct cross-speaker overlap requires at least 60~ms of simultaneous activity, corresponding to two 30~ms VAD frames. Qualifying overlap fragments separated by at most 500~ms are counted as one event without adding the intervening gap to simultaneous-speech duration. The same protocol is applied to reconstruction and expansion.
\end{itemize}

\begin{table}[t]
\centering
\caption{Expansion statistics and interaction analysis. Arrows show paired reconstruction $\rightarrow$ expansion over the evaluated pairs.}
\label{tab:expansion}
\small
\setlength{\tabcolsep}{4pt}
\begin{tabularx}{\columnwidth}{@{}Xr@{}}
\toprule
Metric & Value \\
\midrule
Expansion Factor & 2.340 \\
Turns/min & 19.48 $\rightarrow$ 18.59 \\
Backchannels/min & 2.22 $\rightarrow$ 1.93 \\
Interruptions/min & 5.67 $\rightarrow$ 4.76 \\
Overlap events/min & 9.62 $\rightarrow$ 8.84 \\
\bottomrule
\end{tabularx}
\end{table}

The expansions are 2.340 times as long as their paired reconstructions on average. Their turn rate differs by $-4.6\%$ and their backchannel rate by $-13.2\%$; overlap-event density is $8.0\%$ lower, while interruption density is $16.0\%$ lower. Taken together, these statistics indicate that expansion closely matches reconstruction in turn and overlap-event density, while backchannel and interruption rates remain lower. Because these metrics describe interaction structure rather than semantics, content coherence and dialogue naturalness are evaluated separately.

\subsection{Expansion Content and Dialogue Quality}

Using Gemini's multimodal capabilities, we evaluate how natural each expansion sounds and how naturally its content continues the paired reconstruction. \textbf{Content Coherence} ranges from 1 (unrelated, contradictory, or not a coherent continuation) to 5 (a highly coherent continuation of the established context). \textbf{Dialogue Naturalness} ranges from 1 (not believable as a two-person dialogue) to 5 (highly natural, spontaneous, and believable). The prompt instructs the evaluator to ignore audio fidelity, recording quality, speaker-identity similarity, accent, and annotation accuracy.

\begin{table}[H]
\centering
\caption{Audio-based Gemini judgments for paired reconstruction and expansion. Higher is better; each evaluated pair receives equal weight.}
\label{tab:expansion-content}
\small
\begin{tabular}{lc}
\toprule
Metric & Mean Score \\
\midrule
Content Coherence & 4.940 \\
Dialogue Naturalness & 4.800 \\
\bottomrule
\end{tabular}
\end{table}

Gemini assigns mean scores of 4.94 for content coherence and 4.80 for dialogue naturalness. Its rationales identify direct topical continuation throughout the evaluated pairs and describe the exchanges as natural back-and-forth conversations. Thus, within this automatic evaluation, expansion preserves conversational context and produces plausible dialogue content.

\subsection{Audio-tag Annotation Quality}

Each reconstruction and expansion utterance containing at least one declared audio tag is evaluated independently. Gemini receives the utterance waveform, tagged transcript, and declared tags, and judges only whether the tagged behavior is acoustically expressed at the appropriate position and in the specified order. The five-point rubric ranges from 1 (not expressed at all) to 5 (perfectly expressed). Untagged utterances are excluded from the mean.

\begin{table}[H]
\centering
\caption{Audio-tag Alignment Score. Higher is better; 5 denotes perfect acoustic expression of the declared tags.}
\label{tab:audio-tags}
\small
\begin{tabular}{lc}
\toprule
Stage & Alignment Score \\
\midrule
Reconstruction & 4.376 \\
Expansion & 3.750 \\
Overall & 4.221 \\
\bottomrule
\end{tabular}
\end{table}

Table~\ref{tab:audio-tags} shows stronger correspondence between declared annotations and acoustic realizations in reconstruction than in the regenerated expansion. Reconstruction scores 4.376 and expansion scores 3.750, yielding an overall Audio-tag Alignment Score of 4.221 out of 5. Expansion therefore leaves more room to improve expressive delivery control while retaining the other benefits measured above.

\section{Discussion}

\subsection{Why reconstruction is a distinct data operation}

Reconstruction is not ordinary enhancement. Enhancement changes a waveform while aiming to retain the observed performance. \system instead produces a new waveform from an immutable transcript, a stable identity anchor, and utterance-local delivery evidence. This design can remove residual separation artifacts and standardize single-speaker track quality. More importantly, it makes supervision explicit: the generated waveform is linked to plain text, tagged text, instruction, and aligned words by construction.

Reconstruction improves NISQA and DNSMOS over its paired separated source and retains a positive identity margin. Source-utterance-relative matching identifies substantial preservation of turns and overlaps and strong preservation of backchannels.

\subsection{Why grounded expansion matters}

Expansion addresses a limitation that separation and reconstruction share: both are bounded by the content already present in the recording. Grounded generation can increase lexical coverage while conditioning each speaker on source-derived voice evidence and keeping scene context attached to the source. The reported results show strong predicted acoustic quality, a positive mean speaker-discrimination margin, and, in the paired-audio Gemini evaluation, coherent and natural continuation. Expansion closely matches reconstruction in turn and overlap-event density; its backchannel and interruption rates are modestly lower. Its audio-tag alignment also trails reconstruction, leaving expressive delivery control as the clearest area for improvement.

The value of expansion extends beyond its acoustic scores. A full-duplex model needs examples of listening while speaking, rapid feedback, held silence, overlap onset, and interruption recovery. The expansion schema exposes these events as types and placements rather than leaving them implicit in a mixed waveform. The present results indicate that explicit interaction controls can preserve a broadly similar interaction profile while the generated content remains coherent and natural.

\subsection{A layered view of data fidelity}

The three outputs offer different notions of fidelity and should not be reduced to a single ranking. Separation is faithful to the recorded waveform and naturally observed timing. Reconstruction is content and interaction faithful: it preserves the source words and relative conversational organization while replacing the acoustic rendering with cleaner, explicitly aligned supervision. Expansion is context faithful: it introduces new content while constraining identity, persona, scene, and interaction style. The measured duration and interaction-rate differences are compatible with these different objectives and are best interpreted alongside the additional control and coverage each generated stage provides.

This layering supports a staged training mixture. Separation can provide naturally occurring timing and interaction evidence; reconstruction can add clean, content-aligned examples with explicit expressive supervision; expansion can then broaden the range of grounded conversational events. The transcript schema stays compatible across the generated stages, simplifying batching and objective design. This staged mixture is enabled by the data representation but has not been evaluated as a training curriculum. A downstream study could evaluate it by comparing matched model configurations trained on separation alone, separation plus reconstruction, and all three stages.

\section{Limitations, Ethics, and Release Considerations}

The evaluation is automatic. NISQA and DNSMOS predict perceptual judgments under their training conditions and may react differently to synthesized speech. WER includes error from the remote ASR model; for expansion, it measures adherence to a generated script rather than semantic plausibility. Because the canonical transcript is generated automatically rather than manually annotated, the separation and reconstruction WER values measure agreement with the pipeline's lexical reference rather than absolute transcription accuracy. Speaker cosine similarity can remain high when local pronunciation or emotion is wrong, and it may be partially coupled to reference construction. The interaction detector is deterministic and frozen for the run, but its calibration status is unverified. Expansion content and dialogue quality are judged by one Gemini model on English evaluation data; the high scores are descriptive rather than an estimate of corpus-wide human preference. The Audio-tag Alignment Score evaluates only declared tags; it does not search for unannotated audible events and therefore does not measure annotation recall. The study also does not measure downstream full-duplex model performance.

Real-world audio introduces rights and privacy obligations. Processing and releasing a recording requires a lawful basis, respect for source licenses and platform terms, and safeguards for personally identifying or sensitive speech. Voice cloning creates additional misuse risk because identity can be reproduced beyond the source utterance. A responsible release should document provenance, restrict disallowed sources, evaluate memorization and impersonation risk, provide a removal process, and distinguish released metadata from audio that cannot legally be redistributed. Generated expansions should be clearly labeled as synthetic and not presented as statements actually made by the original speakers.

Bias can enter through source selection, diarization, ASR, quality filters, language models, and TTS. Filters may preferentially retain studio-like voices and common language varieties, while rejecting accents, dialects, noisy environments, and overlapping styles that are important for robust dialogue modeling. Separate reporting by language, accent, gender presentation, acoustic condition, and source domain is needed where such analysis is lawful and ethically appropriate.

Finally, the pipeline composes third-party models and services whose versions, licenses, and behavior can change. A release should pin model revisions when possible, retain configuration and hashes, and record which stage used a remote service. These records are necessary for reproducibility and for honoring the conditions attached to every component.

\section{Conclusion}

\system converts real monaural conversations into three linked forms of full-duplex speech data with complementary training roles. Quality-verified separation recovers speaker-specific streams, canonical text, and naturally observed interaction timing. Source-faithful reconstruction regenerates the same exchange as cleaner, word-aligned speech with explicit expressive supervision. Conversation-grounded expansion adds new interactions constrained by the recovered speakers, context, and interaction pattern. The generated stages retain a common artifact schema with equal-duration single-speaker tracks, word-aligned transcripts, audio-tag anchors, utterance types, and delivery instructions.

The automatic evaluation supports this division of labor. All stages retain positive speaker-discrimination margins and strong predicted acoustic quality. Reconstruction preserves backchannels strongly and retains substantial turn and overlap correspondence despite the duration change introduced by resynthesis. Expansion receives mean scores of 4.94 for coherence with reconstruction and 4.80 for two-person dialogue naturalness, and its interaction rates remain close to reconstruction: turn and overlap-event density differ by 4.6\% and 8.0\%, while backchannel and interruption rates are 13.2\% and 16.0\% lower. Audio-tag alignment is stronger in reconstruction than expansion, but the overall score remains 4.221 out of 5. Taken together, these differences are consistent with distinct roles for the three generated artifacts. Separation offers natural interaction evidence, reconstruction offers cleaner and more controllable supervision, and expansion offers broader grounded conversational coverage. The results do not establish downstream effectiveness, which requires matched training studies and independent perceptual validation. Within these boundaries, the three-stage pipeline provides a practical route from abundant but entangled real-world audio to progressively structured supervision for models designed to learn when to speak, when to listen, and how conversation unfolds between words.

\bibliographystyle{unsrtnat}
\footnotesize
\bibliography{references}

@article{nguyen2022dgslm,
  title         = {Generative Spoken Dialogue Language Modeling},
  author        = {Nguyen, Tu Anh and Kharitonov, Eugene and Copet, Jade and Adi, Yossi and Hsu, Wei-Ning and Elkahky, Ali and Tomasello, Paden and Algayres, Robin and Sagot, Benoit and Mohamed, Abdelrahman and Dupoux, Emmanuel},
  journal       = {Transactions of the Association for Computational Linguistics},
  volume        = {11},
  pages         = {250--266},
  year          = {2023},
  doi           = {10.1162/tacl_a_00545},
  url           = {https://arxiv.org/abs/2203.16502}
}

@article{defossez2024moshi,
  title         = {Moshi: A Speech-Text Foundation Model for Real-Time Dialogue},
  author        = {D{\'e}fossez, Alexandre and Mazar{\'e}, Laurent and Orsini, Manu and Royer, Am{\'e}lie and P{\'e}rez, Patrick and J{\'e}gou, Herv{\'e} and Grave, Edouard and Zeghidour, Neil},
  journal       = {arXiv preprint arXiv:2410.00037},
  year          = {2024},
  url           = {https://arxiv.org/abs/2410.00037}
}

@inproceedings{lin2025fullbench,
  title         = {{Full-Duplex-Bench}: A Benchmark to Evaluate Full-Duplex Spoken Dialogue Models on Turn-Taking Capabilities},
  author        = {Lin, Guan-Ting and Lian, Jiachen and Li, Tingle and Wang, Qirui and Anumanchipalli, Gopala and Liu, Alexander H. and Lee, Hung-yi},
  booktitle     = {Proceedings of the IEEE Automatic Speech Recognition and Understanding Workshop (ASRU)},
  year          = {2025},
  url           = {https://arxiv.org/abs/2503.04721}
}

@article{nakata2026duplexchat,
  title         = {{DuplexChat}: Constructing Speaker-Separated Full-Duplex Dialogue Speech at Scale for Spoken Dialogue Language Modeling},
  author        = {Nakata, Wataru and Saito, Yuki and Saruwatari, Hiroshi},
  journal       = {arXiv preprint arXiv:2607.04941},
  year          = {2026},
  url           = {https://arxiv.org/abs/2607.04941}
}

@article{nakata2026dialoguesidon,
  title         = {{DialogueSidon}: Recovering Full-Duplex Dialogue Tracks from In-the-Wild Dialogue Audio},
  author        = {Nakata, Wataru and Saito, Yuki and Yamauchi, Kazuki and Tsunoo, Emiru and Saruwatari, Hiroshi},
  journal       = {arXiv preprint arXiv:2604.09344},
  year          = {2026},
  url           = {https://arxiv.org/abs/2604.09344}
}

@inproceedings{subakan2021sepformer,
  title         = {Attention Is All You Need in Speech Separation},
  author        = {Subakan, Cem and Ravanelli, Mirco and Cornell, Samuele and Bronzi, Mirko and Zhong, Jianyuan},
  booktitle     = {Proceedings of the IEEE International Conference on Acoustics, Speech and Signal Processing (ICASSP)},
  pages         = {21--25},
  year          = {2021},
  doi           = {10.1109/ICASSP39728.2021.9413901},
  url           = {https://arxiv.org/abs/2010.13154}
}

@article{chen2022wavlm,
  title         = {{WavLM}: Large-Scale Self-Supervised Pre-Training for Full Stack Speech Processing},
  author        = {Chen, Sanyuan and Wang, Chengyi and Chen, Zhengyang and Wu, Yu and Liu, Shujie and Chen, Zhuo and Li, Jinyu and Kanda, Naoyuki and Yoshioka, Takuya and Xiao, Xiong and others},
  journal       = {IEEE Journal of Selected Topics in Signal Processing},
  volume        = {16},
  number        = {6},
  pages         = {1505--1518},
  year          = {2022},
  doi           = {10.1109/JSTSP.2022.3188113},
  url           = {https://arxiv.org/abs/2110.13900}
}

@inproceedings{mittag2021nisqa,
  title         = {{NISQA}: A Deep {CNN}-Self-Attention Model for Multidimensional Speech Quality Prediction with Crowdsourced Datasets},
  author        = {Mittag, Gabriel and Naderi, Babak and Chehadi, Assmaa and M{\"o}ller, Sebastian},
  booktitle     = {Proceedings of Interspeech},
  pages         = {2127--2131},
  year          = {2021},
  doi           = {10.21437/Interspeech.2021-299},
  url           = {https://arxiv.org/abs/2104.09494}
}

@inproceedings{reddy2021dnsmos,
  title         = {{DNSMOS}: A Non-Intrusive Perceptual Objective Speech Quality Metric to Evaluate Noise Suppressors},
  author        = {Reddy, Chandan K. A. and Gopal, Vishak and Cutler, Ross},
  booktitle     = {Proceedings of the IEEE International Conference on Acoustics, Speech and Signal Processing (ICASSP)},
  pages         = {6493--6497},
  year          = {2021},
  doi           = {10.1109/ICASSP39728.2021.9414878},
  url           = {https://arxiv.org/abs/2010.15258}
}

@article{du2025naturalvoices,
  title         = {{NaturalVoices}: A Large-Scale, Spontaneous and Emotional Podcast Dataset for Voice Conversion},
  author        = {Du, Zongyang and Chandra, Shreeram Suresh and Ulgen, Ismail Rasim and Mahapatra, Aurosweta and Salman, Ali N. and Busso, Carlos and Sisman, Berrak},
  journal       = {arXiv preprint arXiv:2511.00256},
  year          = {2025},
  url           = {https://arxiv.org/abs/2511.00256}
}

@article{xie2025soulx,
  title         = {{SoulX-Podcast}: Towards Realistic Long-Form Podcasts with Dialectal and Paralinguistic Diversity},
  author        = {Xie, Hanke and Lin, Haopeng and Cao, Wenxiao and Guo, Dake and Tian, Wenjie and Wu, Jun and Wen, Hanlin and Shang, Ruixuan and Liu, Hongmei and Jiang, Zhiqi and others},
  journal       = {arXiv preprint arXiv:2510.23541},
  year          = {2025},
  url           = {https://arxiv.org/abs/2510.23541}
}

@article{hu2026qwen3tts,
  title         = {{Qwen3-TTS} Technical Report},
  author        = {Hu, Hangrui and Zhu, Xinfa and He, Ting and Guo, Dake and Zhang, Bin and Wang, Xiong and Guo, Zhifang and Jiang, Ziyue and Hao, Hongkun and Guo, Zishan and others},
  journal       = {arXiv preprint arXiv:2601.15621},
  year          = {2026},
  url           = {https://arxiv.org/abs/2601.15621}
}

\end{document}